\documentclass[11pt]{article}

\usepackage[T1]{fontenc}
\usepackage[utf8]{inputenc}
\usepackage{lmodern}       
\usepackage{microtype}       

\usepackage{graphicx}%
\usepackage{multirow}%
\usepackage{amsmath,amssymb,amsfonts}%
\usepackage{amsthm}%
\usepackage{mathrsfs}%
\usepackage[title]{appendix}%
\usepackage{xcolor}%
\usepackage{textcomp}%
\usepackage{manyfoot}%
\usepackage{booktabs}%
\usepackage{algorithm}%
\usepackage{algorithmicx}%
\usepackage{algpseudocode}%
\usepackage{listings}
\usepackage{subcaption}
\usepackage{pifont}
\usepackage{natbib}
\usepackage{tikz}
\usetikzlibrary{positioning, arrows, shapes.geometric}
\newcommand{\tikzvae}{
\begin{tikzpicture}
  \node (x) at (0, 0) [draw, circle, fill=gray!30, minimum size=1cm] {$\mathbf{x}$};
  \node (z) at (2, 0) [draw, circle] {$\mathbf{z}$};
  \draw[->, dashed] (x) -- node[above] {$q(\mathbf{z}|\mathbf{x})$} (z);
  
  \node (x') at (4, 0) [draw, circle, fill=gray!30, minimum size=1cm] {$\mathbf{x'}$};
  \draw[->] (z) -- node[above] {$p(\mathbf{x}|\mathbf{z})$} (x');
  
  \node (pz) at (2, 2) [draw, circle] {$p(\mathbf{z})$};
  \draw[->, dashed] (pz) -- (z);
\end{tikzpicture}
}

\newcommand{\tikzpivae}{
\begin{tikzpicture}
  \node (x) at (0, 0) [draw, circle, fill=gray!30, minimum size=1cm] {$\mathbf{x}$};
  \node (k) at (2.5, 0) [draw, circle] {$\mathbf{z}_{\text{phy}}$};
  \draw[->, dashed] (x) -- node[above] {$q(\mathbf{z}_{\text{phy}}|\mathbf{x})$} (k);
  
  \node (x') at (6, 0) [draw, circle, fill=gray!30, minimum size=1cm] {$\mathbf{x'}$};
  \draw[->] (k) -- node[above] {$p(\mathbf{x}|\mathbf{z}_{\text{phy}}, x_c)$} (x');
  
  \node (pk) at (2.5, 2) [draw, circle] {$p(\mathbf{z}_{\text{phy}})$};
  \draw[->, dashed] (pk) -- (k);
\end{tikzpicture}
}

\newcommand{\tikzgpvae}{
\begin{tikzpicture}
  \node (x) at (-2, 0) [draw, circle, fill=gray!30, minimum size=1cm] {$\mathbf{x}$};
  \node (z) at (0, 0) [draw, circle] {$\mathbf{z}$};
  \node (x') at (2, 0) [draw, circle, fill=gray!30, minimum size=1cm] {$\mathbf{x'}$};
  
  \draw[->, dashed] (x) -- node[above] {$q(\mathbf{z}|\mathbf{x})$} (z);

  \draw[->] (z) -- node[above] {$p(\mathbf{x}|\mathbf{z})$} (x');
  
  \node (pz) at (0, 2) [draw, circle] {$p(\mathbf{z} |\mathbf{t})$};
  \draw[->, dashed] (pz) -- (z);
\end{tikzpicture}
}

\newcommand{\tikzpigpvae}{
\begin{tikzpicture}
  \node (x) at (0, 0) [draw, circle,fill=gray!30] {$\mathbf{x}$};
  
  \node (z_phy) at (2.8, 1) [draw, circle] {$\mathbf{z}_{\text{phy}}$};
  \node (z_delta) at (2.8, -1) [draw, circle] {$\mathbf{z}_{\delta}$};
  
  \node (x') at (5.6, 0) [draw, circle,fill=gray!30] {$\mathbf{x'}$};
  
  \draw[->, dashed] (x) -- node[above left] {$q(\mathbf{z}_{\text{phy}}|\mathbf{x})$} (z_phy);
  \draw[->, dashed] (x) -- node[below left] {$q(\mathbf{z}_{\delta}|\mathbf{x})$} (z_delta);
  
  \draw[->] (z_phy) -- node[above right] {$p(\mathbf{x}|\mathbf{z}_{\text{phy}})$} (x');
  \draw[->] (z_delta) -- node[below right] {$p(\mathbf{x}|\mathbf{z}_{\delta})$} (x');
  
  \node (pz_phy) at (2.8, 3) [draw, circle] {$p(\mathbf{z}_{\text{phy}})$};
  \node (pz_delta) at (2.8, -2.85) [draw, circle] {$p(\mathbf{z}_{\delta}|\mathbf{t})$};
  
  \draw[->, dashed] (pz_phy) -- (z_phy);
  \draw[->, dashed] (pz_delta) -- (z_delta);
\end{tikzpicture}
}

\graphicspath{{figures/}}

\definecolor{darkgreen}{RGB}{0,150,0}  % Define a darker green
\definecolor{darkred}{RGB}{180,0,0} % Define a darker red
\newcommand{\tickg}{\textcolor{darkgreen}{\ding{52}}}  % Define the green tick
\newcommand{\tickr}{\textcolor{darkred}{\ding{52}}}

\usepackage{hyperref}
\hypersetup{
    colorlinks=true,
    linkcolor=blue,
    citecolor=blue,
    urlcolor=blue,
    pdfauthor={Your Name},
    pdftitle={Paper title},
    pdfkeywords={keyword1, keyword2}
}
\makeatletter
\renewcommand{\@fnsymbol}[1]{\@arabic{#1}}
\makeatother

\title{PIGPVAE: Physics-Informed Gaussian Process Variational Autoencoders}
\author{
  Michail Spitieris\thanks{SINTEF Digital} \and
  Abdulmajid Murad\footnotemark[1] \and
  Massimiliano Ruocco\thanks{SINTEF Digital and Norwegian University of Science and Technology} \and
  Alessandro Nocente\thanks{SINTEF Community} \\
  \footnotesize\texttt{\{michail.spitieris,abdulmajid.murad,massimiliano.ruocco,alessandro.nocente\}@sintef.no}
}

\begin{document}

\maketitle

\abstract{Recent advances in generative AI offer promising solutions for synthetic data generation but often rely on large datasets for effective training. To address this limitation, we propose a novel generative model that learns from limited data by incorporating physical constraints to enhance performance. Specifically, we extend the VAE architecture by incorporating physical models in the generative process, enabling it to capture underlying dynamics more effectively. While physical models provide valuable insights, they struggle to capture complex temporal dependencies present in real-world data. To bridge this gap, we introduce a discrepancy term to account for unmodeled dynamics, represented within a latent Gaussian Process VAE (GPVAE). Furthermore, we apply regularization to ensure the generated data aligns closely with observed data, enhancing both the diversity and accuracy of the synthetic samples. The proposed method is applied to indoor temperature data, achieving state-of-the-art performance. Additionally, we demonstrate that PIGPVAE can produce realistic samples beyond the observed distribution, highlighting its robustness and usefulness under distribution shifts.}

\section{Introduction}

Recent advances in generative AI methods have demonstrated remarkable success across domains such as image and text generation, producing outputs often indistinguishable from real data \citep{rombach2022high}.  
However, their application to areas like indoor temperature modeling, where data collection is costly and limited, remains challenging due to the reliance on substantial amounts of training data.
Considering the high costs of collecting indoor temperature data, this work aims to develop generative models that can learn effectively from limited datasets

Optimal control of HVAC systems presents a significant challenge, particularly in accurately forecasting indoor temperatures, which is crucial for balancing energy efficiency. 
A major obstacle is the limited availability of high-quality data, as acquiring such data is both costly and energy-intensive. Indoor temperature data exhibits strong temporal dependencies, introducing an additional layer of complexity to the modeling process. 
Recent approaches for time series data have primarily adapted existing architectures such as Generative Adversarial Networks (GANs) \citep{gans2014}, Variational Autoencoders (VAEs) \citep{vaes2013}, Normalizing flows (NFs) \citep{papamakarios2021normalizing}, and diffusion models (DMs) \citep{ho2020denoising}, modifying them to account for temporal dynamics. 
Examples include TimeGAN \citep{timeGAN}, TimeVQVAE \citep{timeVQVAE}, and Fourier flows \citep{alaa2021generative}. See also \citep{tsgBench} for a review and comparison.
Another line of work utilizes Gaussian Process (GP) priors to model the temporal dependencies in the latent space of the VAE model \citep{pearce_gaussian_2020,jazbec_scalable_2021, tran_fully_2023} and has achieved state-of-the-art performance in various tasks.

Some methods generate samples that closely resemble the training data—so closely that they lack diversity—while others are difficult to train and tend to produce samples that diverge from the data distribution. 
Our goal is to develop generative methods capable of producing sufficiently diverse time series samples that also preserve the data distribution.
Additionally, these models must be capable of learning from small sample sizes (e.g., 20 data points in our case).

In scenarios where data is scarce, incorporating prior knowledge about the data generation process can significantly enhance the learning process. 
In engineering disciplines, such prior knowledge is often available through physical models, which provide valuable constraints and improve explainability. 
A notable example of this approach is the successful application of physical models in physics-informed machine learning, where the integration of physical laws has led to more accurate and interpretable results \citep{PIML}. 

While physical models are essential to design and control HVAC system, accurately capturing room temperature dynamics as a consequence of the HVAC system action is a much more complex problem.
Real-world complexities such as uneven heat distribution, thermal inertia and heat accumulation in the materials, variable occupant behaviour, and changing environmental conditions makes capturing room temperature dynamics very challenging.
Consequently, physical models often require supplementation with empirical data and data-driven models to effectively reflect the actual thermal behaviour of a room.
For instance, heating and cooling processes tend to follow approximately exponential curves, which can be modelled using Newton's law of cooling (or heating).

In this work, we adopt the VAE architecture due to its simplicity and robustness during training, modifying it to incorporate physical constraints in both the generator and the latent space. 
While the physical model alone cannot fully capture all the temporal dependencies present in the data, it is capable of modeling the main trends. 
To address the limitations of the physical model, we introduce a discrepancy term that accounts for the missing physics in a data-driven manner. 
Specifically, we draw inspiration from the GPVAE architecture to probabilistically learn and generate these missing dynamics, ensuring that the model reflects both the known physical laws and the unmodeled complexities in the data.

A potential drawback of the discrepancy component is its flexibility, which could dominate the model and obscure the underlying physics. 
To mitigate this, we incorporate a regularization term that minimizes the discrepancy, encouraging the physical model to align closely with the full model.
This ensures a balanced representation of both the physical processes and the missing dynamics, enhancing the model’s interpretability and performance.

In summary, the proposed method combines traditional approaches for simulating scenarios in science and engineering through physical models with the powerful data-driven capabilities and representation learning capacity of modern deep generative methods.

The paper is organized as follows: Section \ref{sec:related_work} reviews related work. Section \ref{sec:proposed_models} introduces the proposed model. In Section \ref{sec:Experiments}, we present the experimental results and compare the proposed method with state-of-the-art approaches. Finally, Section \ref{sec:conlcusion} discusses the advantages and limitations of the proposed generative model, along with potential future research directions.

\section{ Related Work} \label{sec:related_work}
Generative models for synthetic time series generation adapt existing models like GANs \citep{gans2014}, VAEs \citep{vaes2013}, and flow-based models \citep{papamakarios2021normalizing} to account for temporal dependencies.
For example, TimeGAN \citep{timeGAN} combines supervised and unsupervised learning by embedding time series data into a latent space to capture temporal dynamics and uses adversarial training. 
Motivated by the success of the VQ-VAE model \citep{vqvae2017}, timeVQVAE \cite{timeVQVAE} decomposes time series into low and high frequencies using STFT for enhanced time series modeling.

\noindent \textbf{Variational Autoencoders} In this work, we build on the VAE model due to its simplicity and stability during training. 
VAEs use an encoder network which parameterizes a posterior distribution \( q(\mathbf{z}|\mathbf{x}) \) of the latent variables \( \mathbf{z} \), a prior \( p(\mathbf{z}) \), and a decoder model with a distribution \( p(\mathbf{x}|\mathbf{z}) \) (see Fig. \ref{fig:standard_vae}).
The model is trained by maximizing the Evidence Lower Bound (ELBO) given by:
\begin{equation}
    \mathcal{L}_{\text{VAE}}(\theta, \phi; \mathbf{x}) = \mathbb{E}_{q_\phi(\mathbf{z} \mid \mathbf{x})}\left[\log p_\theta(\mathbf{x} \mid \mathbf{z})\right] - \text{KL}\left(q_\phi(\mathbf{z} \mid \mathbf{x}) \,\|\, p_\theta(\mathbf{z})\right) \label{eq:VAE}
\end{equation}
The first term of Eq. \eqref{eq:VAE} is the reconstruction loss, where with continuous data the mean squared error (MSE) is used:
\begin{equation}
    -\mathbb{E}_{q_\phi(\mathbf{z} \mid \mathbf{x})}\left[\log p_\theta(\mathbf{x} \mid \mathbf{z})\right] \propto \|\mathbf{x} - \hat{\mathbf{x}}\|^2 = \frac{1}{n} \sum_{i=1}^{n} (\mathbf{x}_i - \hat{\mathbf{x}}_i)^2    
\end{equation}
and the second term is the KL divergence between the posterior distribution \( q_\phi(\mathbf{z} \mid \mathbf{x}) \) and the prior \( p_\theta(\mathbf{z}) \). Independent Gaussian priors are commonly used due to their analytical tractability, and result in the following equation:
\begin{equation}
    \text{KL}\left(q_\phi(\mathbf{z} \mid \mathbf{x}) \,\|\, p_\theta(\mathbf{z})\right) = \frac{1}{2} \sum_{l=1}^{L} \left( \sigma_l^2(\mathbf{x}) + \mu_l^2(\mathbf{x}) - 1 - \log \sigma_l^2(\mathbf{x}) \right)
\end{equation}
where $L$ is the dimension of the latent variables.

\vspace{0.1cm}
\noindent \textbf{Gaussian process VAE} The Gaussian process VAE (GPVAE) \citep{casale2018GPVAE, pearce2020GPVAE} model drops the independence assumption and models the temporal dependencies in the latent variables through a GP prior (see Fig. \ref{fig:gpvae}).
Specifically, the model in \cite{pearce2020GPVAE} performs exact GP posterior inference.
For each latent dimension \( l=1,\ldots, L \), the inputs of the GP model are the conditional inputs \( \mathbf{t} \) (time in our case), while the outputs \( \tilde{\mathbf{x}}_l=\mu^l_\phi(\mathbf{x}) \) are generated by the inference network \( q^* \) with heteroscedastic noise with \( \tilde{\sigma}_l=\sigma^l_\phi(\mathbf{x}) \).
More specifically, the GP prior is \( \mathbf{z} \sim GP(0, K(\mathbf{t}, \mathbf{t} \mid \theta)) \), and the ELBO is given by the following expression (see also \ref{appsec:gpvae} for derivation):
\begin{equation}
    \mathcal{L}_{\text{GPVAE}} = 
    \mathbb{E}_{q(\mathbf{z} \mid \cdot)} 
    \left[ \log p_{\psi} (\mathbf{x} \mid \mathbf{z}) - \log \tilde{q}_{\phi}(\mathbf{z} \mid \mathbf{x}) \right] +  \log Z_{\phi, \theta} (\mathbf{x}, \mathbf{t})
\end{equation}
\noindent \textbf{Accounting for missing physics} Physics-informed Machine Learning (PIML) \citep{PIML2021physics} is an emerging field that integrates physical knowledge into machine learning algorithms. However, the effectiveness of PIML depends on the accuracy of the underlying physical models, which are often constrained by simplifying assumptions. A growing body of work addresses this limitation by jointly modeling the missing physics alongside the known physical models \citep{koh2001bayesian, marmin2022deep, spitieris2023bayesian}, with a primary focus on solving inverse problems and improving predictive accuracy. Similar ideas have been applied in generative modeling contexts \citep{takeishi_physics-integrated_2021}, primarily focusing on reconstruction and forecasting tasks rather than on synthetic data generation.
\begin{figure}[H]
\centering
\begin{minipage}{0.48\textwidth}
\centering
\tikzvae
\caption{Standard VAE}
\label{fig:standard_vae}
\end{minipage}
\hfill
\begin{minipage}{0.48\textwidth}
\centering
\tikzgpvae
\caption{GPVAE}
\label{fig:gpvae}
\end{minipage}
\end{figure}

\section{Proposed Generative Model} \label{sec:proposed_models}
In VAEs temporal dynamics can be accounted in two ways:\\
1. \textbf{Temporal Dynamics in the Decoder Model}:
Many VAE variants employ decoders that can capture temporal dependencies in the data. 
A straightforward example is using Recurrent Neural Networks (RNNs) or Long Short-Term Memory (LSTM) networks within the decoder to model time dependencies.
In Sec. \ref{sec:PIVAE},  we introduce a VAE model which models the temporal dynamics using a physical model as a decoder.\\
2. \textbf{Temporal Dynamics in the Latent Space}: An alternative approach is to model temporal dependencies directly in the latent space, where the VAE learns to encode these dynamics. A prominent example is the GPVAE (Gaussian Process VAE) model described in Sec. \ref{sec:related_work}.
In Sec. \ref{sec:PIGPVAE}, we introduce a VAE model that combines the PIVAE with the GPVAE architectures in a way that the GPVAE accounts for the missing physics of the PIVAE decoder.

\subsection{PIVAE} \label{sec:PIVAE}

The physics-informed VAE (PIVAE) replaces the standard neural network decoder
$p(\mathbf{x}\mid \mathbf{z})$ with a physical model. 
This can be a physical model with a known solution, a model solver, or even a physics-informed neural network (PINN) \citep{pinns2019} introducing additional constraints in the objective function.
For example, the Newton's law of cooling (or heating) can simulate how temperature decreases (or increases) through time, and is described by the following first-order differential equation
\begin{equation} 
\dfrac{dT}{dt} = -k (T - T_s), \label{eq:NL} 
\end{equation} 
where $T$ is the temperature, $T_s$ is the temperature of the surroundings, and $k$ is a constant that controls the rate of temperature increase (heating) or decrease (cooling) over time.
The solution to Eq. \eqref{eq:NL} is
\begin{equation} 
T(t) = (T_0 - T_s) \exp(-k \cdot t) + T_s, \label{eq:NL_sol} 
\end{equation} 
where $T_0$ is the initial temperature.
In this equation, the only unknown parameter is $k$, which can be estimated from the training data. The VAE framework allows estimating a distribution over $k$, as it can be treated as a latent variable replacing $\mathbf{z}$ in the traditional VAE architecture (see Fig. \ref{fig:standard_vae} and \ref{fig:pivae}).
Note that is possible to use any physical model.
The PIVAE model offers several advantages:
\textbf{Interpretable Latent Space} the latent variable $k$ has a clear physical interpretation, making the latent space more meaningful.
\textbf{Parsimonious Parametrization} by replacing the neural network decoder with the physical equation \eqref{eq:NL_sol}, the model has no training parameters.
This is particularly helpful in  cases with limited training data since it avoids overfitting.
The generative process in PIVAE is conditional on time $t$, $T_0,$ $T_s$ and the latent parameter $k$. 
The conditioning variables are denoted as $\mathbf{x}_c = (t, T_0, T_s).$

In the general case where where we have a physical model with physical parameters $\mathbf{z}_\text{phy}$ and conditioning variables $\mathbf{x}_c,$ the objective function (ELBO) for the PIVAE model is given as
\begin{equation} 
\mathcal{L}_\text{PIVAE} = \mathbb{E}_{q_\psi(\mathbf{z}_\text{phy} \mid \cdot)} \left[ \log p(\mathbf{x} \mid \mathbf{z}_\text{phy}, \mathbf{x}_c) \right] - \text{KL}\left( q_\psi(\mathbf{z}_\text{phy} \mid \mathbf{x} )  \,\|\,  p(\mathbf{z}_\text{phy}) \right). 
\end{equation}
This formulation encourages the model to learn a meaningful distribution over $\mathbf{z}_\text{phy}$, while ensuring that the generated data follows the dynamics descibed by the physical model.

In this context, we typically have prior knowledge about the distribution of $k$. 
Unlike the traditional VAE model in Eq. \eqref{eq:VAE}, which uses a prior with zero mean and unit variance, the prior for $\mathbf{z}_\text{phy}$ in PIVAE has a defined mean and variance. 
The KL divergence between the approximate posterior and this prior is given by:
\begin{equation} 
\text{KL}\left( \mathcal{N}(\mu_q, \sigma_q^2) \,\|\, \mathcal{N}(\mu_p, \sigma_p^2) \right) = \log \left( \frac{\sigma_p}{\sigma_q} \right) + \frac{\sigma_q^2 + (\mu_q - \mu_p)^2}{2\sigma_p^2} - \frac{1}{2}. \label{eq:KL_PIVAE} 
\end{equation}
This formulation reflects how the PIVAE leverages prior knowledge of $\mathbf{z}_\text{phy}$ to improve model interpretability and performance, compared to the traditional VAE approach.
\begin{figure}[htbp]
\begin{minipage}{0.48\textwidth}
\centering
\tikzpivae
\caption{PIVAE}
\label{fig:pivae}
\end{minipage}
\hfill
\begin{minipage}{0.48\textwidth}
\centering
\tikzpigpvae
\caption{PIGPVAE}
\label{fig:pigpvae}
\end{minipage}
\end{figure}

\subsection{PIGPVAE} \label{sec:PIGPVAE}

The PIVAE model in Sec. \ref{sec:PIVAE} employees a physical model as a decoder, which is typically a simplified representation of the complex underlying physics. 
To achieve a richer representation we introduce another term in the PIVAE model to capture the missing physics.
The decoder network of the PIVAE takes a sample $\mathbf{z}_{\text{phy}}$ and along with the conditioning variables of the physical model $\mathbf{x}_c$ they are fed to the physical model to reconstruct the observed data.
Let us denote for simplicity the decoder of the PIVAE model as $f_{\text{phy}}(\mathbf{z}_{\text{phy}}, \mathbf{x}_c),$ and its evaluation for a specific sample $\tilde{\mathbf{z}}_{\text{phy}},$ $\hat{\mathbf{x}}_{\text{phy}} = f_{\text{phy}}(\tilde{\mathbf{z}}_{\text{phy}}, \mathbf{x}_c).$
We introduce the missing physics decoder model $f_{\delta},$ which can be a NN as in the GPVAE model. 
This model is a function of the latent variable $\mathbf{z}_\delta\sim GP(0,K(\mathbf{t}, \mathbf{t})),$ along with $\hat{\mathbf{x}}_{\text{phy}},$ $f_{\delta}(\mathbf{z}_\delta,\hat{\mathbf{x}}_{\text{phy}}).$
For given samples $\tilde{\mathbf{z}}_{\text{phy}},\tilde{\mathbf{z}}_{\delta}$ and $\hat{\mathbf{x}}_{\text{phy}} = f_{\text{phy}}(\tilde{\mathbf{z}}_{\text{phy}}, \mathbf{x}_c),$ the decoder of the PIGPVAE model is the sum of the physical model decoder with the discrepancy model as follows
\begin{equation}
      f_{\text{phy}}(\tilde{\mathbf{z}}_{\text{phy}}, \mathbf{x}_c) + f_{\delta}(\tilde{\mathbf{z}}_\delta,\hat{\mathbf{x}}_{\text{phy}}).
\end{equation}
See also Fig. \ref{fig:pigpvae} for a graphical representation of the model. 
By considering the prior decomposition $p(\mathbf{z}_{\text{phy}},\mathbf{z}_\delta) = p(\mathbf{z}_\delta \mid \mathbf{z}_{\text{phy}}) p(\mathbf{z}_{\text{phy}}),$ the ELBO is given by the following expression
\begin{multline}
    \mathcal{L}_\text{PIGPVAE} = \mathbb{E}_q \left[\log p(\mathbf{x} \mid \mathbf{z}_{\text{phy}}, \mathbf{z}_\delta) 
     -\log q^*(\mathbf{z}_\delta \mid \mathbf{z}_{\text{phy}}, \mathbf{x}) \right] + \log Z(\mathbf{x}, \mathbf{t}) - \\
    \text{KL}(q(\mathbf{z}_{\text{phy}} \mid \mathbf{x}) \| p(\mathbf{z}_{\text{phy}})) \label{eq:PIGPVAE}.
\end{multline}
Eq. \eqref{eq:PIGPVAE}, is a combination of the PIVAE and GPVAE ELBOs (see Appendix for the derivation).

A potential issue with this model is the excessive flexibility of the discrepancy decoder which can dominate the model and produce estimates of the physical parameters $\mathbf{z}_\text{phy}$ that are not faithful to the physical model.
To penalize such flexibility we introduce a regularization term which encourage the (decoder) physical model to generate samples
close to the observed data.
More specifically, the MSE loss between the full model $\mathbf{\hat{x}}$ and the physical model $\hat{\mathbf{x}}_{\text{phy}}$ generated data is minimized. This is equivalent to minimizing the discrepancy between the physical model and the observed data.
The regularized loss function is given by 
\begin{multline}
    \mathcal{L}_\text{PIGPVAE} = 
    \mathbb{E}_q \left[\log p(\mathbf{x} \mid \mathbf{z}_{\text{phy}}, \mathbf{z}_\delta) 
    - \log q^*(\mathbf{z}_\delta \mid \mathbf{z}_{\text{phy}}, \mathbf{x}) \right] + \log Z(\mathbf{x}) \\
    - \text{KL}(q(\mathbf{z}_{\text{phy}} \mid \mathbf{x}) \| p(\mathbf{z}_{\text{phy}})) 
    + \alpha\|\mathbf{x} - \mathbf{x}_\text{phy}\|^2. \label{eq:PIGPVAE_reg}
\end{multline}
The $\alpha$ parameter controls the effect of the regularization. 
Larger values of $\alpha$ enforce greater fidelity to the physics-based reconstructions, making the outputs closely adhere to the physical model.
While for smaller values of alpha the model becomes more flexible allowing for more realistic reconstructions. 
As $\alpha \to 0$, the discrepancy model's influence dominates, potentially overshadowing the physical model.

\begin{figure}[H]
    \centering
    \includegraphics[width=\textwidth]{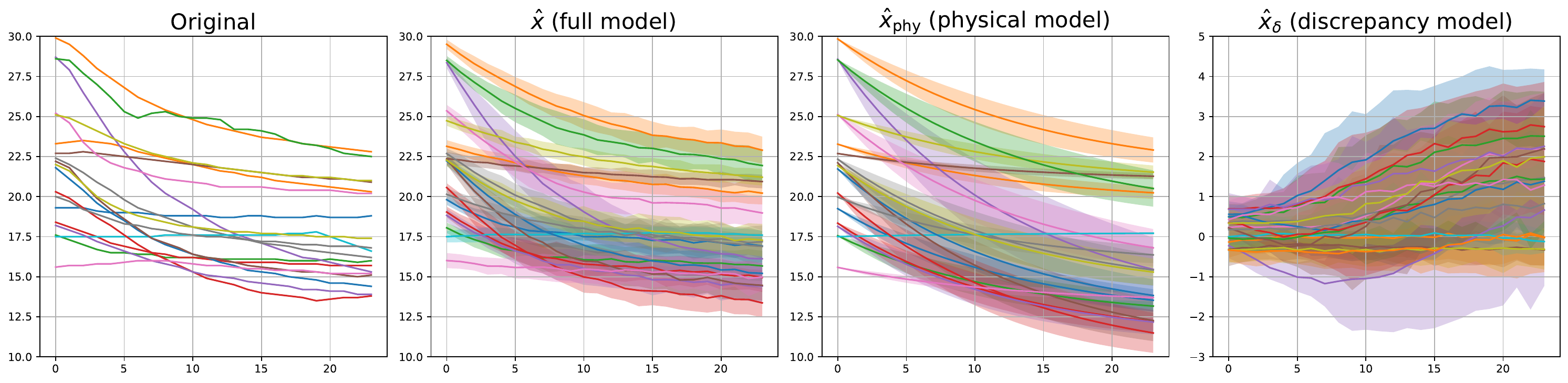}
    \caption{PIGPVAE model; Reconstructed cooling data. Left plot shows the original data and $\hat{\mathbf{x}} $ (full model) is the PIGPVAE model reconstructions which are a combination of the physical model a nd the learned discrepancies.}
    \label{fig:reconstructed_quantities}
\end{figure}

In the implementation described in Sec. \ref{sec:Experiments}, $\alpha$ is treated as a trainable parameter and optimized jointly with the other model parameters to achieve the best balance between physical faithfulness and reconstruction flexibility.
A graphical representation of the model is given in Fig. \ref{fig:reconstructed_quantities}, where the original cooling data (left) are reconstructed by the full model, which is the addition of the physical model (Newton's Law), $\hat{\mathbf{x}}_\text{phy}$ with the discrepancy,  $\hat{\mathbf{x}}_\delta$ (GPVAE model).

\section{Experiments} \label{sec:Experiments}
In this experimental study, we use data from the RICO dataset \citep{rico_data} \footnote{dataset is available at https://doi.datacite.org/dois/10.60609\%2Ftw79-4k72 and code will be publicly available after the review process}, specifically comprising 29 heating and 28 cooling curves. For the PIVAE and PIGPVAE models, we use the temperature from the cooling and heating systems (ventilation and radiation heating/cooling) as the surrounding temperature $T_s$. Each curve consists of 24 time steps. The generative models are trained separately on the heating and cooling data, using $70\%$ of the available data for training.

To evaluate and compare performance, we train five generative models: GPVAE, PIVAE, PIGPVAE, TimeVQVAE, and TimeGAN. The latter two models are included due to their promising results on various benchmark datasets \citep{tsgBench}. For the TimeGAN model, we use the implementation from \cite{nikitin2024tsgm} and for TimeVQVAE we use the original implementation \citep{timeVQVAE}.

The performance of the models is evaluated using both visual and quantitative metrics. Visualization includes plots of the generated samples alongside the original data, density plots comparing the original and synthetic data distributions, and dimensionality reduction techniques (PCA and t-SNE) to examine structural similarities between the generated and real data. For quantitative metrics, we use Maximum Mean Discrepancy (MMD) and Marginal Distribution Difference (MDD) due to their robustness \citep{tsgBench}. We exclude model-based metrics like Discriminative Score (DS) \citep{timeGAN} and Predictive Score (PS) \citep{timeGAN}, as the small sample size can produce inconsistent results.

The optimal control of HVAC systems depends on the initial room temperature before applying heating or cooling. However, the experimental data do not cover all potential initial temperature scenarios. To address this limitation, we consider two case studies: 1) \textbf{In-distribution generation}, where the domain of generation matches the observed data (see Sec. \ref{sec:exp_in}), and 2) \textbf{Out-of-distribution generation}, where the domain of generation extends beyond the observed range (see Sec. \ref{sec:exp_out}). For the latter, we define a cutoff value of $20^\circ\mathrm{C}$ and remove all data below this threshold when training the models for both heating and cooling. The models are then used to generate data with starting values below $20^\circ\mathrm{C}$ (if possible).

\begin{figure}[h!]
    \centering
    \includegraphics[width=\textwidth]{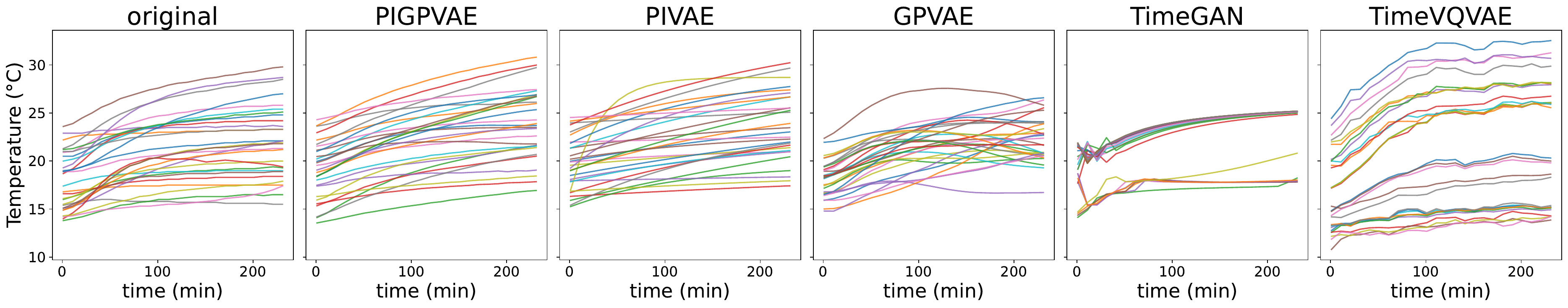}
    
    \vspace{1em}
    
    \includegraphics[width=\textwidth]{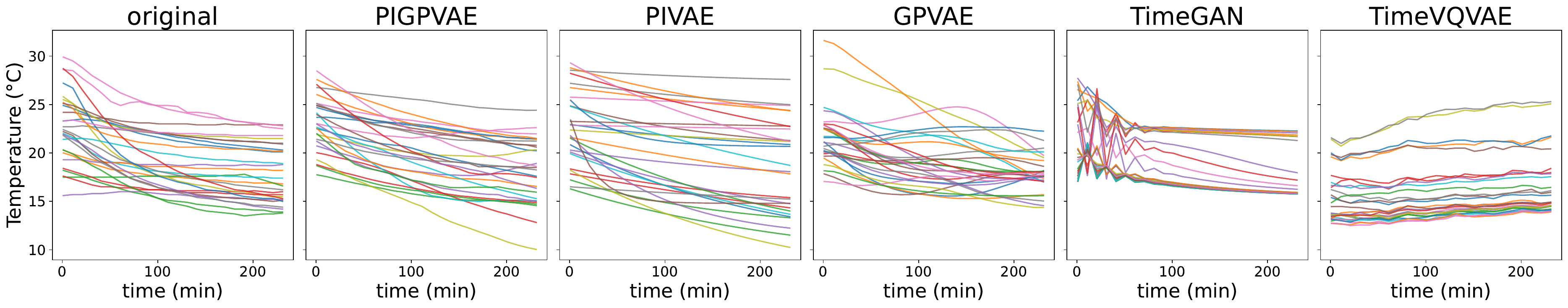}
    \caption{Visual comparison of generated vs real data. The top figure shows heating data, while the bottom figure shows cooling data.}
    \label{fig:gen_vs_orig_in}
\end{figure}
\begin{figure}[h!]
    \centering
    \includegraphics[width=\textwidth]{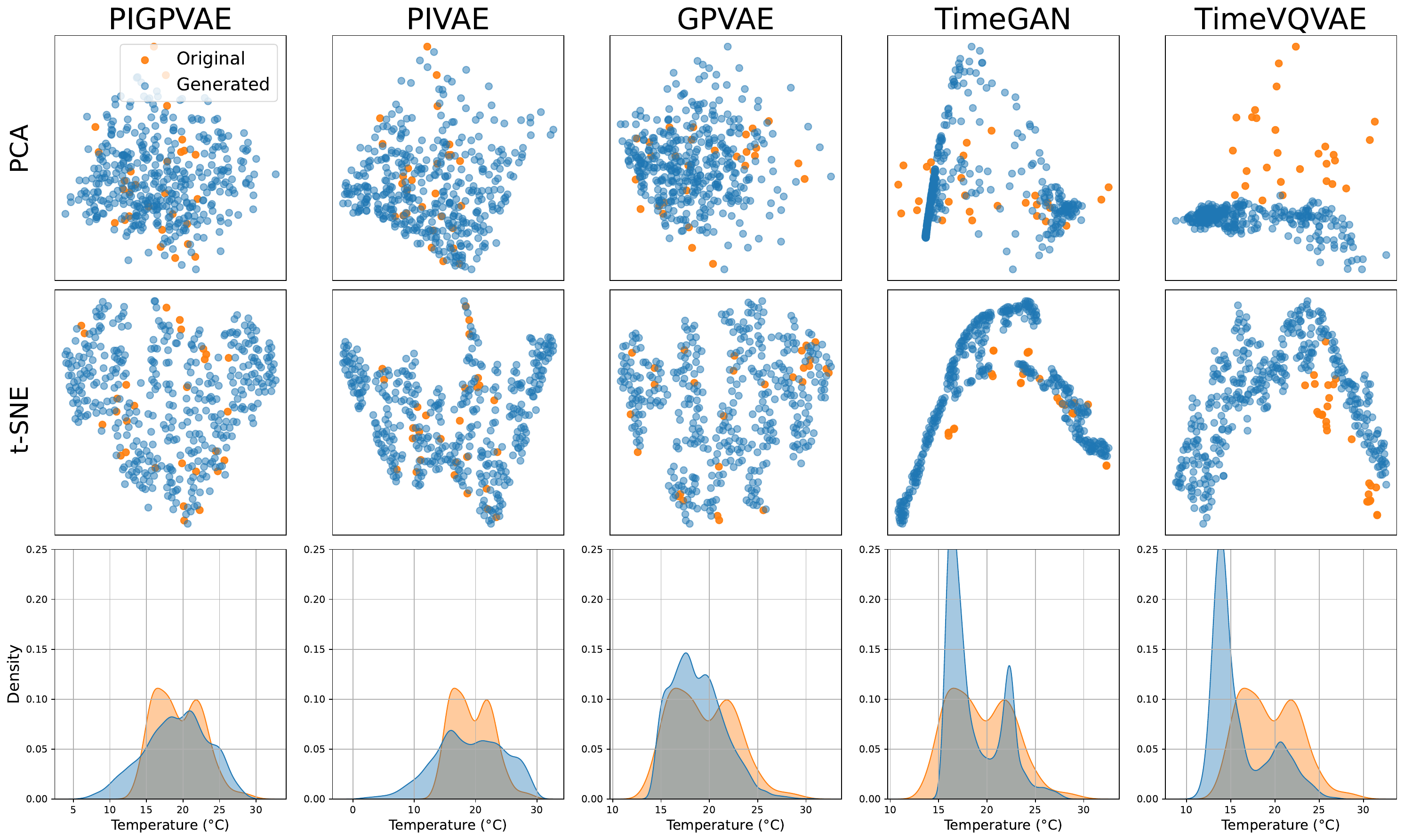} 
    \caption{PCA, t-SNE and densities; cooling data.}
    \label{fig:PCA_t-SNE_cooling_in}
\end{figure}
\begin{figure}[h!]
    \centering
    \includegraphics[width=\textwidth]{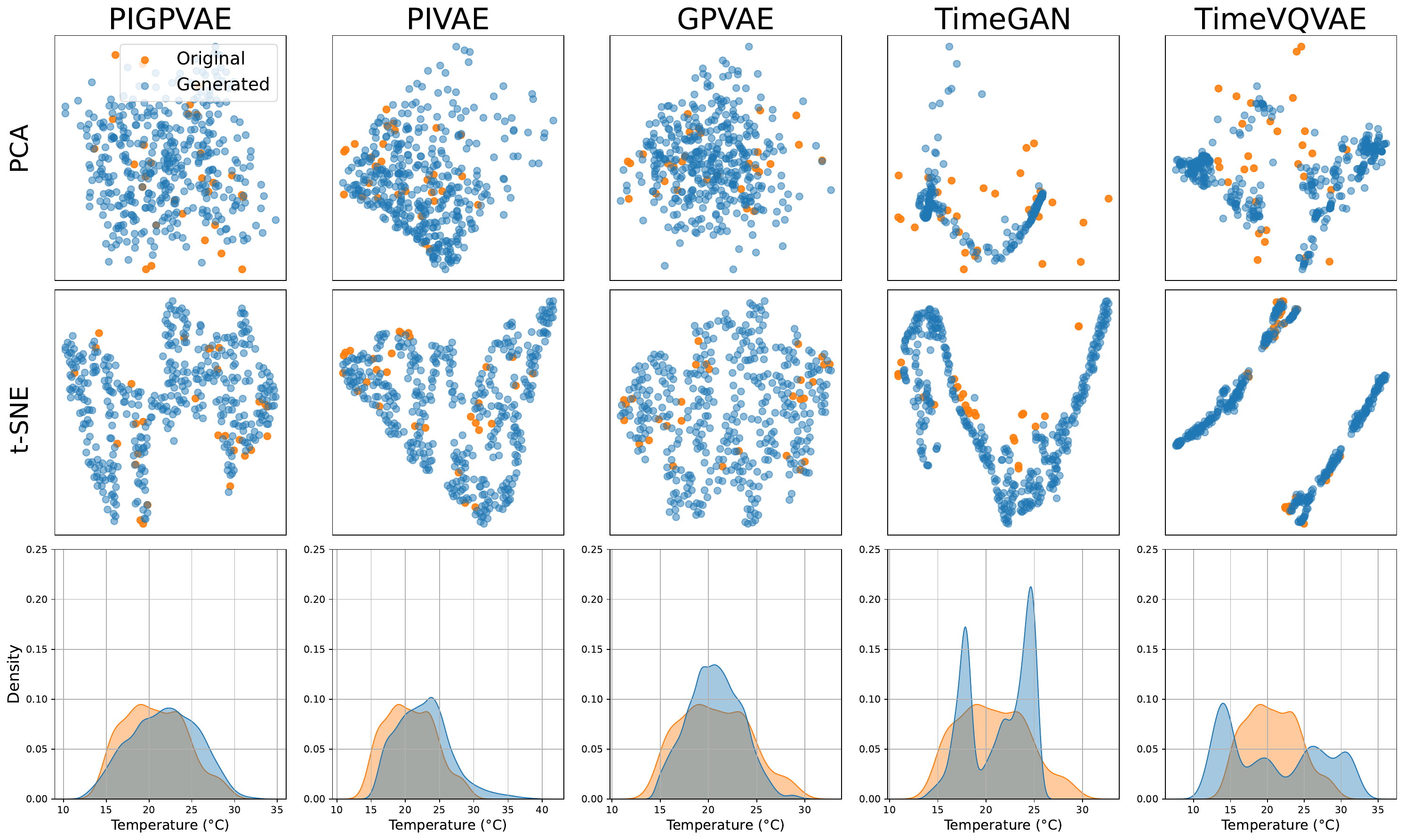} 
    \caption{PCA, t-SNE and densities; heating data.}
    \label{fig:PCA_t-SNE_heating_in}
\end{figure}

\subsection{Metrics}
\textbf{Maximum Mean Discrepancy} (MMD) \citep{MMD2012} measures the similarity between two probability distributions $P$ and $Q$ as follows 
\begin{equation}
    \text{MMD}^{2}(P,Q) = E_{P}~ [k(X,X)] + E_{Q}~ [k(Y,Y)] - 2 E_{P,Q}~ [k(X,Y)].
\end{equation}
In practice we have original and generated samples, MMD can be estimated as follows
\begin{multline}
    \text{MMD}^{2}(X,Y) = 
    \frac{1}{m (m-1)} \sum_{i} \sum_{j \neq i} k(\mathbf{\mathbf{x}_{i}}, \mathbf{\mathbf{x}_{j}})  
    + \frac{1}{m (m-1)} \sum_{i} \sum_{j \neq i} k(\mathbf{y_{i}}, \mathbf{y_{j}})\\
    - 2 \frac{1}{m^2} \sum_{i} \sum_{j} k(\mathbf{\mathbf{x}_{i}}, \mathbf{y_{j}}),
\end{multline}
where $k(\mathbf{\mathbf{x}_{i}}, \mathbf{\mathbf{x}_{j}}) = \exp \left(\frac{- \Vert \mathbf{\mathbf{x}_{i}} - \mathbf{\mathbf{x}_{j}} \Vert^{2}}{2\sigma^{2}}\right).$ \\

\vspace{0.1cm}

\noindent \textbf{Correlation difference}  (CD) \citep{ni2021sig} measures the difference in correlation between the original samples $X$  and generated samples $\hat{X}$.
\begin{equation}
    \text{CD}(X, \hat{X})=\sum_{i,j=1}^n |\rho (\mathbf{x}_i, \mathbf{x}_j) - \rho (\hat{\mathbf{x}}_i, \hat{\mathbf{x}}_j)|
\end{equation}

\vspace{0.1cm}

\noindent \textbf{Marginal Distribution Difference}  (MDD) \citep{tsgBench} measures the difference in distribution at each time step between the original and generated data using the centers of histograms bins.

\subsection{Case 1: In distribution generation} \label{sec:exp_in}
In this experimental study, we aim to generate data that aligns with the observed data distribution. The models are trained on $70\%$ of the observed data and subsequently evaluated on the entire dataset, consisting of 20 data points for heating and 20 data points for cooling.
The models are trained separately on heating and cooling data.

In Fig. \ref{fig:gen_vs_orig_in}, we observe that TimeGAN performs poorly in generating realistic synthetic data. TimeVQVAE shows slightly better performance but fails to generate much variety in both heating and cooling curves. While the GPVAE generates data with greater variety, it does not preserve the underlying dynamics of heating and cooling. For instance, some heating curves increase and then decrease, while some cooling curves decrease and then increase.
In contrast, PIVAE preserves the dynamics of heating and cooling, though the generated data appear overly idealized and do not look realistic. 
Finally, the PIGPVAE combines the best of the two approaches by preserving the dynamics while also capturing intricate details of the original data through the inclusion of model discrepancy.

In Figs. \ref{fig:PCA_t-SNE_cooling_in} and \ref{fig:PCA_t-SNE_heating_in}, we observe that none of the methods perfectly capture the data distribution due to the limited training sample size. 
However, the proposed PIGPVAE method performs the best, as evidenced by the PCA and t-SNE plots and the metrics in Table \ref{tab:In}.
PIGPVAE outperforms other methods across all metrics except the CD (correlation difference) metric for cooling.
This is because the generated samples from the PIVAE are highly correlated and therefor the withing generated data correlation is high.
\begin{table}[h!]
\scriptsize
\centering
\setlength{\tabcolsep}{2pt}
\renewcommand{\arraystretch}{0.9}
\begin{tabular}{llll}
\toprule
 & \multicolumn{3}{c}{Heating} \\
\cmidrule(lr){2-4}
Model & MMD  & CD & MDD  \\
\midrule
TimeGAN & 0.1979 (0.0531) & 188.8527 (55.1017) & 0.0360 (0.0023) \\
TimeVQVAE & 0.0923 (0.0081) & 112.6020 (9.5220) & 0.0445 (0.0057) \\
GPVAE & 0.0702 (0.0022) & 294.1780 (73.1341) & 0.0235 (0.0033) \\
PIVAE & 0.0782 (0.0035) & 95.5722 (2.2680) & 0.0615 (0.0082) \\
PIGPVAE & \textbf{0.0688} (0.0029) & \textbf{95.2850} (6.5340) & \textbf{0.0165} (0.0013) \\
\specialrule{0.2pt}{0pt}{0pt}
p-value &$<1e^{-4}$  \tickg & 0.7 \ding{56} & $<1e^{-6}$ \tickg\\
\midrule
 & \multicolumn{3}{c}{Cooling} \\
\cmidrule(lr){2-4}
Model & MMD  & CD & MDD  \\
\midrule
TimeGAN & 0.1866 (0.0418) & 169.0639 (21.5154) & 0.0298 (0.0036) \\
TimeVQVAE & 0.1265 (0.0159) & 172.8433 (40.9432) & 0.0487 (0.0082) \\
GPVAE & 0.0732 (0.0024) & 287.5684 (80.5038) & 0.0225 (0.0039) \\
PIVAE & 0.0802 (0.0031) & \textbf{89.7458} (4.6724) & 0.0430 (0.0064) \\
PIGPVAE & \textbf{0.0725} (0.0022) & 114.4033 (35.3524) & \textbf{0.0202} (0.0038) \\
\specialrule{0.2pt}{0pt}{0pt}
p-value & 0.03 \tickg & $<1e^{-6}$ \tickr &  $<1e^{-4}$ \tickg\\
\bottomrule
\end{tabular}
\caption{Metrics for Heating and Cooling; In distribution generation.}
\label{tab:In}
\end{table}

\begin{figure}[h!]
    \centering
    \includegraphics[width=\textwidth]{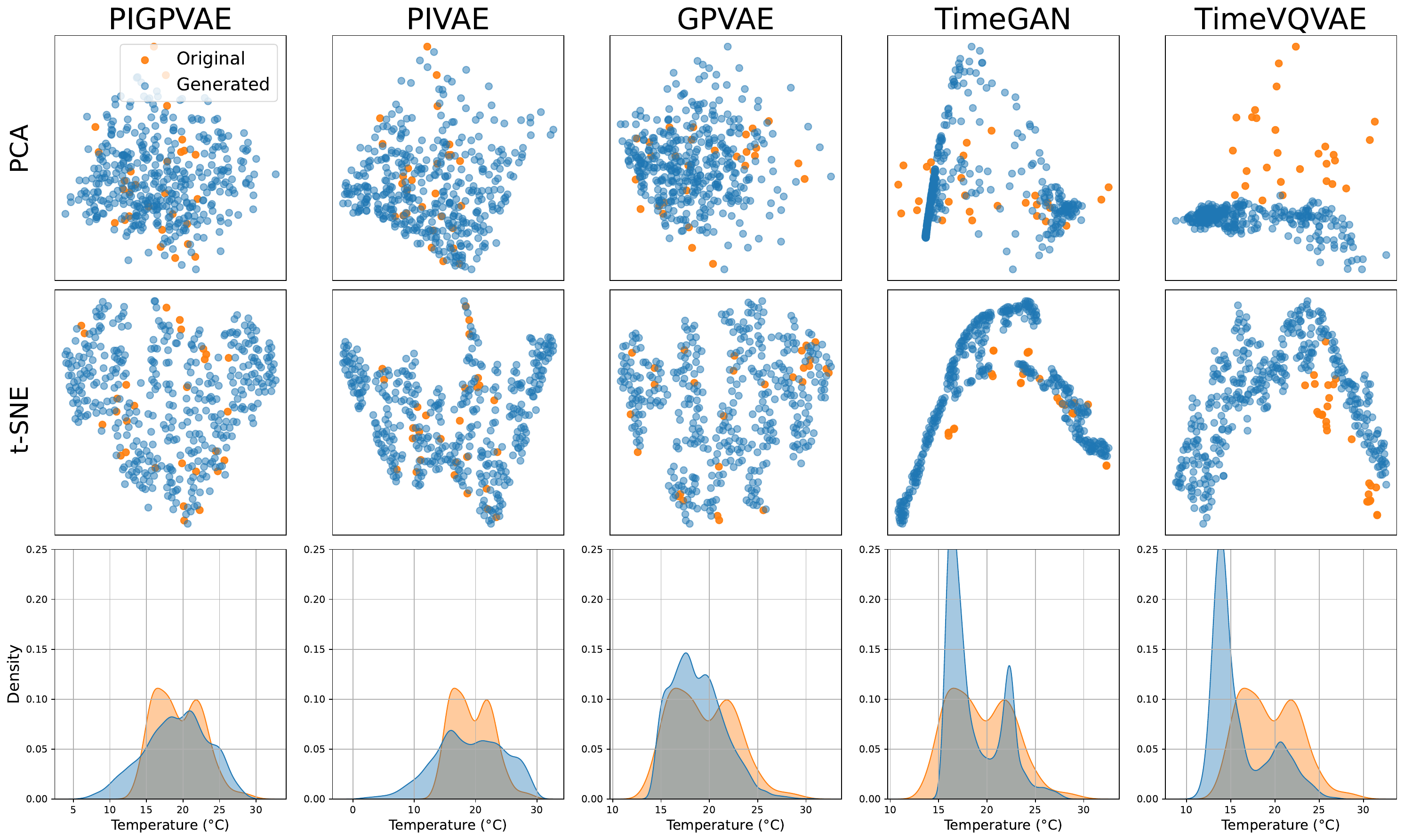} 
    \caption{PCA, t-SNE and densities. In distribution generation; cooling data.}
    \label{fig:PCA_t-SNE_cooling_out}
\end{figure}

\begin{figure}[h!]
    \centering
    \includegraphics[width=\textwidth]{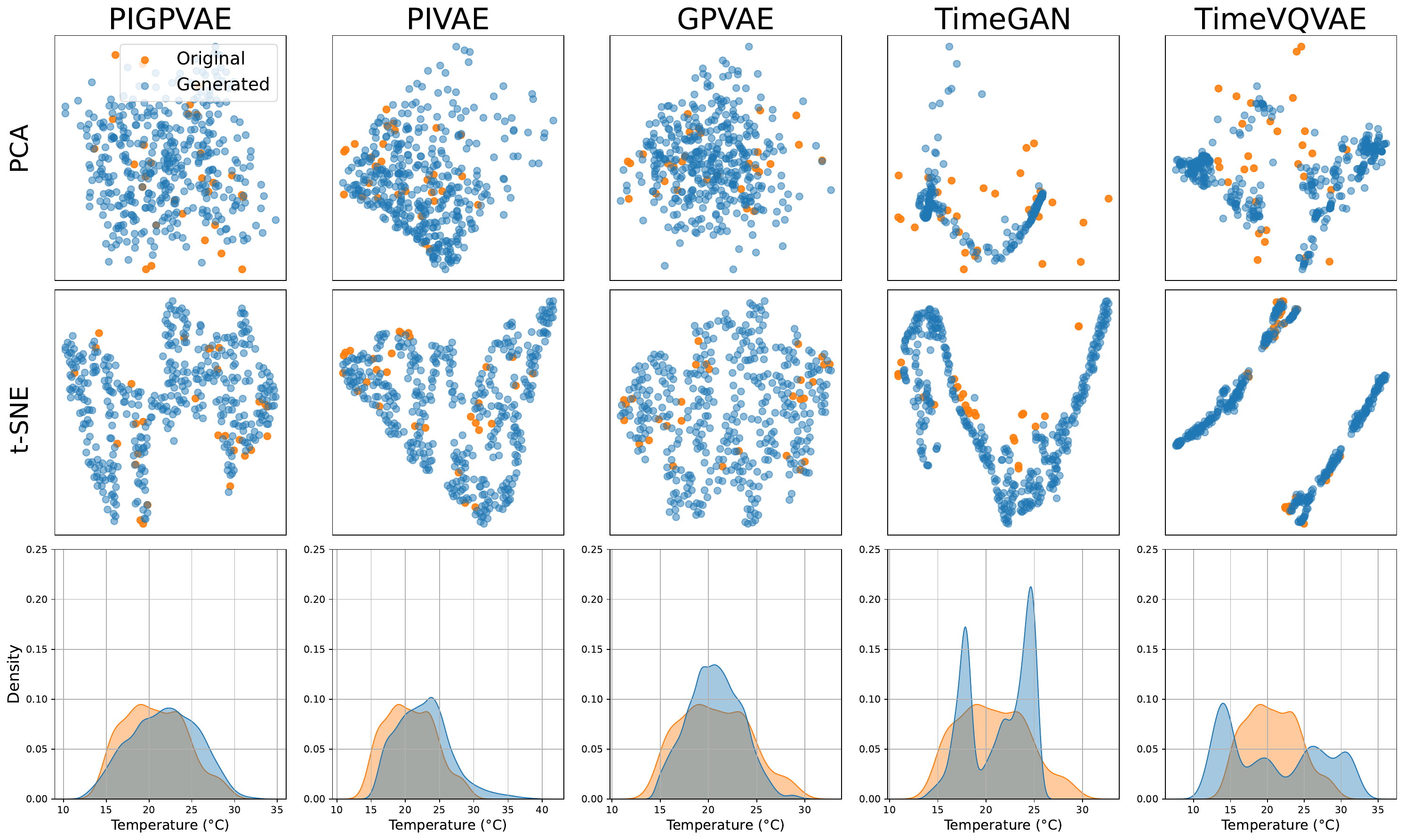} 
    \caption{PCA, t-SNE and densities. In distribution generation;}
    \label{fig:PCA_t-SNE_heating_in}
\end{figure}

\subsection{Case 2: Out of distribution generation} \label{sec:exp_out}

\begin{figure}[h!]
    \centering
    \includegraphics[width=\textwidth]{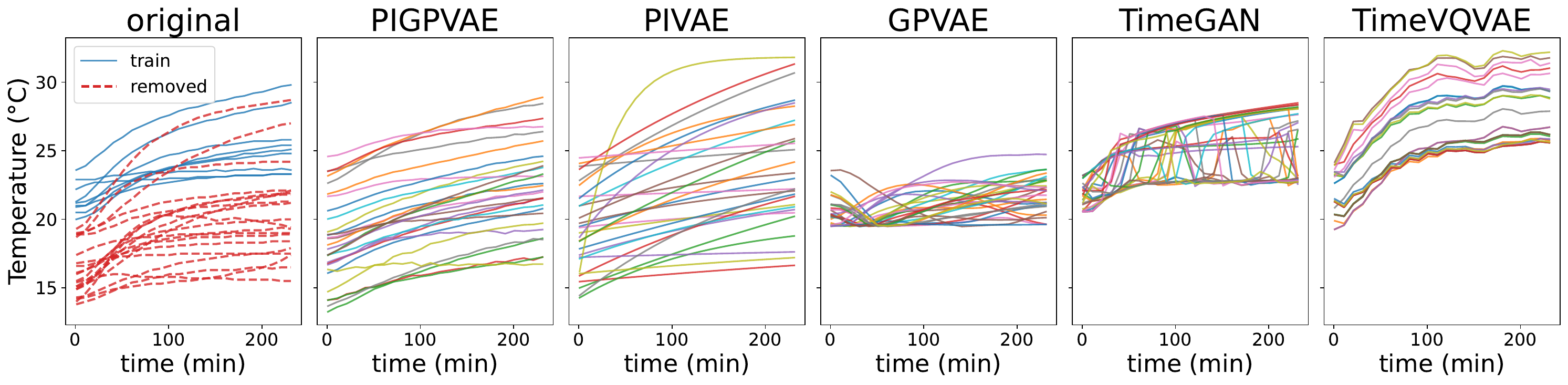}
    
    \vspace{1em}
    
    \includegraphics[width=\textwidth]{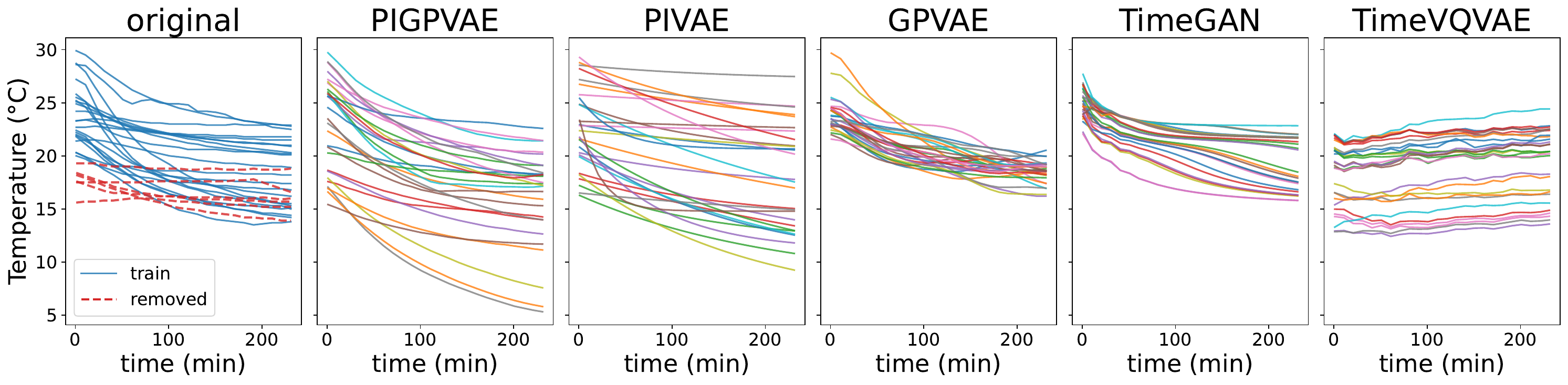}
    \caption{Visual comparison of generated vs real data. The top figure shows heating data (9 train data points), while the bottom figure shows cooling data (20 train data points).}
    \label{fig:gen_vs_orig_out}
\end{figure}

In this experimental study, we consider a scenario where the target domain for data generation is not represented in the observed data. 
Specifically, we remove all samples with starting temperatures below $20^\circ\mathrm{C}$ (see Fig. \ref{fig:gen_vs_orig_out}, left plots). 
The training set includes only $n_\text{tr}=9$ data points for heating models and $n_\text{tr}=20$ for cooling models. The same models as in Sec. \ref{sec:exp_in} are trained.

\begin{figure}[h!]
    \centering
    \includegraphics[width=\textwidth]{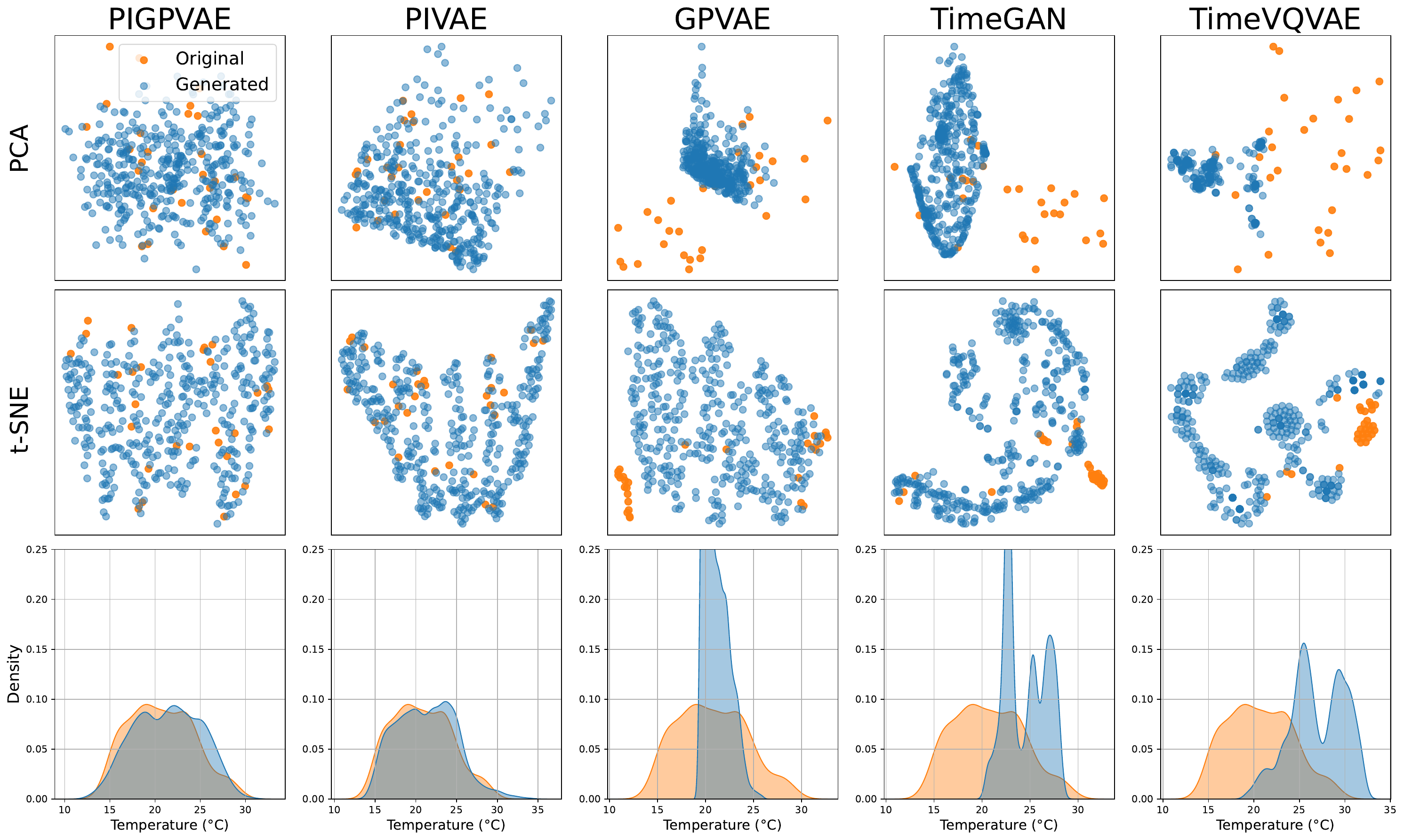} 
    \caption{PCA, t-SNE and densities. Out of distribution generation; cooling data.}
    \label{fig:PCA_t-SNE_heating_out}
\end{figure}

Similar to the in-distribution generation case, TimeGAN performs poorly, while TimeVQVAE shows slight improvement (Fig. \ref{fig:gen_vs_orig_out}). However, fully data-driven models (GPVAE, TimeGAN, and TimeVQVAE) struggle to generate samples outside the observed data distribution, particularly for heating, where much of the domain has been removed (Fig. \ref{fig:gen_vs_orig_out}, top-left). This limitation is also evident in the density plots (Fig. \ref{fig:PCA_t-SNE_cooling_out}, \ref{fig:PCA_t-SNE_heating_out}, bottom row) and the PCA and t-SNE visualizations, where the generated data fail to capture the true distribution.

In contrast, the PIVAE and PIGPVAE models successfully generate out-of-distribution samples by conditioning on starting temperatures outside the observed range. 
The PIGPVAE model, in particular, produces realistic heating and cooling curves by leveraging the physical model decoder’s natural conditioning and learned discrepancies (Fig. \ref{fig:gen_vs_orig_out}). 
Notably, it performs well even with limited data ($n_\text{tr}=9$ for heating). 
Additionally, PIGPVAE effectively covers the original (but unseen) data distribution, as shown in the density plots (Fig. \ref{fig:PCA_t-SNE_cooling_out}, \ref{fig:PCA_t-SNE_heating_out}, bottom row). 
Table \ref{tab:Out} further demonstrates that PIGPVAE outperforms other methods across nearly all metrics.

\section{Conclusion} \label{sec:conlcusion}
In this work, we presented PIGPVAE, a significant advancement in generative modeling by effectively integrating physical constraints with the flexible representation learning capabilities of deep generative models.
The framework is applicable to a broad range of problems where some understanding of the underlying physics exists, but accurately modeling the physics remains challenging.
The PIGPVAE model is interpretable, as it is grounded in physical knowledge that is incorporated during synthetic data generation through the use of physical parameters and conditioning variables. This results in more controlled and reliable data generation.

Through rigorous experimentation, it has been shown to outperform state-of-the-art methods in scenarios with limited data. More importantly, as shown in Sec. \ref{sec:exp_out}, the PIGPVAE model is capable of generating synthetic data beyond the observed data distribution which was not possible with purely data-driven generative models.

A potential limitation is that the model requires some approximate knowledge of the underlying physics, which, while typically available in engineering and scientific applications, may restrict its usability in domains where such prior knowledge is unavailable. 
Additionally, the cubic computational complexity of the latent GP model, $\mathcal{O}(n^3)$, can become prohibitive for modern applications involving long temporal sequences, where scalability is a critical concern. Recent GPVAE variants \citep{jazbec_scalable_2021, fortuin_gp-vae_2020, zhu23b_markovian_GPVAE} address this computational bottleneck through GP approximations, such as sparse GP techniques or structured Markovian assumptions. 
Moreover, a fully Bayesian treatment of the GPVAE model \citep{fully_Bayes_VAE-tran23a} provides a more flexible posterior distribution approximation, enabling improved inference of physical parameters and uncertainty quantification.

Overall, PIGPVAE provides a robust and interpretable framework that holds promise for a range of applications, from anomaly detection to probabilistic imputation, laying the groundwork for future innovations in physics-informed machine learning.
\clearpage
\begin{figure}[h!]
    \centering
    \includegraphics[width=\textwidth]{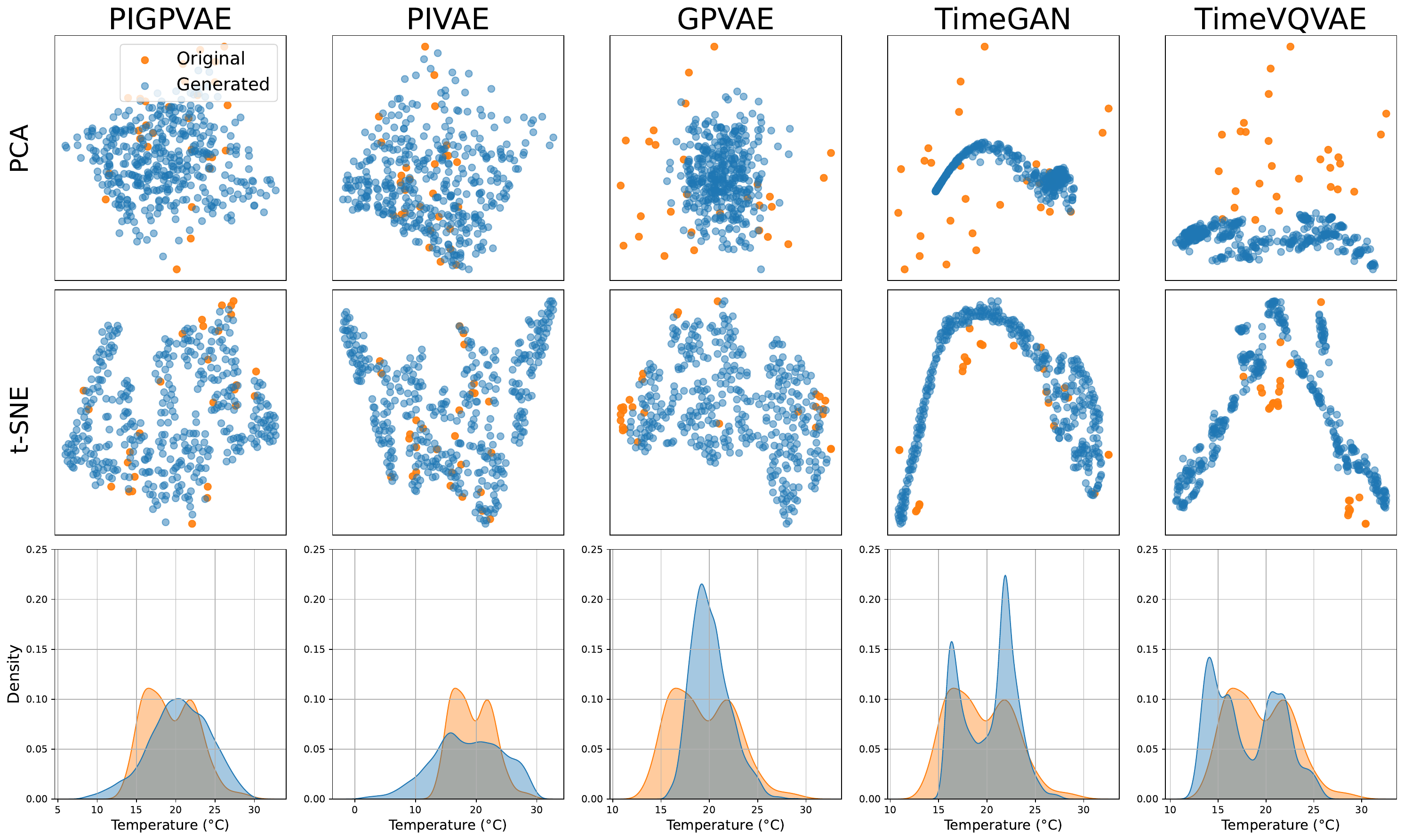} 
    \caption{PCA, t-SNE and densities. Out of distribution generation; heating data.}
    \label{fig:PCA_t-SNE_cooling_out}
\end{figure}
\begin{table}[h!]
\scriptsize
\centering
\setlength{\tabcolsep}{2pt}
\renewcommand{\arraystretch}{0.9}
\begin{tabular}{llll}
\toprule
 & \multicolumn{3}{c}{Heating} \\
\cmidrule(lr){2-4}
Model & MMD  & CD & MDD \\
\midrule
TimeGAN & 0.0772 (0.0042) & 478.3892 (61.7711) & 0.0603 (0.0012) \\
TimeVQVAE & 0.1626 (0.0224) & \textbf{92.3075} (0.5471) & 0.0608 (0.0023) \\
GPVAE & 0.0848 (0.0047) & 504.6981 (76.9627) & 0.0610 (0.0033) \\
PIVAE & 0.0757 (0.0025) & 96.0600 (3.0710) & 0.0571 (0.0082) \\
PIGPVAE & \textbf{0.0733} (0.0042) & 98.1113 (8.7819) &\textbf{0.0259} (0.0056) \\
\specialrule{0.2pt}{0pt}{0pt}
p-value & $<1e^{-5}$ \tickg & $<1e^{-6}$ \tickr &  $<1e^{-6}$ \tickg\\
\midrule
 & \multicolumn{3}{c}{Cooling} \\
\cmidrule(lr){2-4}
Model & MMD  & CD & MDD \\
\midrule
TimeGAN & 0.1513 (0.0280) & 89.7051 (7.8596) & 0.0248 (0.0030) \\
TimeVQVAE & 0.1084 (0.0114) & 325.8537 (42.1474) & 0.0336 (0.0063) \\
GPVAE & 0.0823 (0.0036) & 106.4338 (16.5407) & 0.0365 (0.0031) \\
PIVAE & 0.0798 (0.0032) & 89.9382 (4.5465) & 0.0476 (0.0088) \\
PIGPVAE & \textbf{0.0745} (0.0024) & \textbf{86.7729} (2.0691) & \textbf{0.0206} (0.0037) \\
\specialrule{0.2pt}{0pt}{0pt}
p-value & $<1e^{-6}$ \tickg & $<1e^{-6}$ \tickg &  $<1e^{-6}$ \tickg\\
\bottomrule
\end{tabular}
\caption{Metrics for Heating and Cooling; Out of distribution generation.}
\label{tab:Out}
\end{table}

\clearpage
\bibliography{References}
\clearpage
\begin{appendices}

\section{Objective functions}\label{secA1}
\subsection{VAE}
The marginal likelihood of the data is given by:
\begin{equation}
    \log p_\theta(\mathbf{x}) = \log \int p_\theta(\mathbf{x}, \mathbf{z}) \, d\mathbf{z}
\end{equation}
We introduce the variational approximation to the posterior \( q_\phi(\mathbf{z}|\mathbf{x}) \) as follows:
\begin{align}
    \log p_\theta(\mathbf{x}) &= \log \int \frac{p_\theta(\mathbf{x}, \mathbf{z})}{q_\phi(\mathbf{z}|\mathbf{x})} q_\phi(\mathbf{z}|\mathbf{x}) \, d\mathbf{z}\\
    &= \log \mathbb{E}_{q_\phi(\mathbf{z}|\mathbf{x})} \frac{p_\theta(\mathbf{x}, \mathbf{z})}{q_\phi(\mathbf{z}|\mathbf{x})}
\end{align}
By applying Jensen's inequality, we have that:
\begin{align}
    \log p_\theta(\mathbf{x}) 
    & \geq \mathbb{E}_{q_\phi(\mathbf{z}|\mathbf{x})} \left[ \log \frac{p_\theta(\mathbf{x}, \mathbf{z})}{q_\phi(\mathbf{z}|\mathbf{x})} \right] \\
    &= \mathbb{E}_{q_\phi(\mathbf{z}|\mathbf{x})} \left[ \log p_\theta(\mathbf{x}, \mathbf{z}) - \log q_\phi(\mathbf{z}|\mathbf{x}) \right] \\
    &= \mathbb{E}_{q_\phi(\mathbf{z}|\mathbf{x})} \left[ \log p_\theta(\mathbf{x}|\mathbf{z}) + \log p_\theta(\mathbf{z}) - \log q_\phi(\mathbf{z}|\mathbf{x}) \right] \\
    &= \mathbb{E}_{q_\phi(\mathbf{z}|\mathbf{x})} \left[ \log p_\theta(\mathbf{x}|\mathbf{z}) \right] - \text{KL}(q_\phi(\mathbf{z}|\mathbf{x}) \| p_\theta(\mathbf{z}))
\end{align}

Finally, the objective function to optimize is:
\begin{equation}
    \mathcal{L}(\theta, \phi; \mathbf{x}) = \mathbb{E}_{q_\phi(\mathbf{z}|\mathbf{x})} \left[ \log p_\theta(\mathbf{x}|\mathbf{z}) \right] - \text{KL}(q_\phi(\mathbf{z}|\mathbf{x}) \| p_\theta(\mathbf{z}))
\end{equation}

\subsection{PIVAE}
The objective of the PIVAE model is similar to the VAE model above with two main modifications. 
The decoder is a not a NN anymore, but the Newton's Law of cooling (or heating) given by the following differential equation
\begin{equation}
    \dfrac{dT}{dt} = -k (T - T_s),
\end{equation}
where the solution is given by the following exponential function $(T_0 - T_s)  \exp(-k\cdot t) + T_s.$
The model is a conditional generator, specifically on the initial temperature $T_0$ and the surrounding temperature $T_s.$
The second modification is that we now have one latent variable $k$ instead of $\mathbf{z},$ which defines the rate of change of temperature through time.
The PIVAE objective is given by 
\begin{equation}
    \mathcal{L}(\theta, \phi; T) = \mathbb{E}_{q_\phi(k|T)} \left[ \log p_\theta(T|k,t,T_0, T_s) \right] - \text{KL}(q_\phi(k|T) \| p_\theta(k)). 
\end{equation}
In the general case, where we have a physical model with physical parameters $\mathbf{z}_\text{phy}$ and conditioning variables $\mathbf{x}_c$ the PIVAE ELBO is  
\begin{equation}
    \mathcal{L}(\theta, \phi; \mathbf{x}) = \mathbb{E}_{q_\phi(\mathbf{z}_\text{phy}|\mathbf{x})} \left[ \log p_\theta(\mathbf{x}|\mathbf{x}_c) \right] - \text{KL}(q_\phi(\mathbf{z}_\text{phy}|\mathbf{x}) \| p_\theta(\mathbf{z}_\text{phy})). 
\end{equation}

\subsection{GPVAE} \label{appsec:gpvae}
The GPVAE model captures the temporal dependency in the latent space with a GP prior. 
It can be viewed as a generalization of the VAE. The traditional VAE model assumes that there is no dependency in the latent space $z \sim \mathcal{N}(0,\sigma^2I),$ while the GPVAE captures the temporal dependency in the covariance function $K$, $z \sim GP(0,K(t,t')).$
The decoder model is a NN as in the traditional VAE model.
This model is also a conditional generator since the latent samples are conditioned in time through the covariance function.
\begin{align}
\log p(\mathbf{x}) &= \log \int_{\mathbf{z}} p(\mathbf{z}, \mathbf{x}) \, d\mathbf{z} \\
&= \log \int_{\mathbf{z}} \frac{q(\mathbf{x} \mid \mathbf{z})}{q(\mathbf{x} \mid \mathbf{z})} p(\mathbf{z}, \mathbf{x}) \, d\mathbf{z} \\
&\geq \int_{\mathbf{z}} q(\mathbf{z} \mid \mathbf{x}) \log \frac{p(\mathbf{z}, \mathbf{x})}{q(\mathbf{z} \mid \mathbf{x})} \, d\mathbf{z} \\
&= \int_{\mathbf{z}} q(\mathbf{z} \mid \mathbf{x}) \log p(\mathbf{x} \mid \mathbf{z}) + \log \frac{p(\mathbf{z})}{q(\mathbf{z} \mid \mathbf{x})} \, d\mathbf{z} \\
&= \mathbb{E}_q [\log p(\mathbf{x} \mid \mathbf{z})] + \mathbb{E}_q \left[\log \frac{p(\mathbf{z})}{q(\mathbf{z} \mid \mathbf{x})} \right]\label{eq:gpvae_exp}
\end{align}

This looks identical to the standard VAE model. \cite{pearce_gaussian_2020} suggested replacing the approximate posterior \(q\) with the following approximation:

\begin{equation}
q(\mathbf{x}\mid \mathbf{z}) = \frac{p(\mathbf{z})q^*(\mathbf{z}\mid \mathbf{x})}{Z(\mathbf{x})}.
\end{equation}

By substituting into the second term of Eq.~\eqref{eq:gpvae_exp}, we have:

\begin{align}
    \mathbb{E}_q \left[\log \frac{p(\mathbf{z})}{q(\mathbf{z} \mid \mathbf{x})} \right] &=
    \mathbb{E}_q \left[\log \frac{p(\mathbf{z})}{\frac{p(\mathbf{z})q^*(\mathbf{z} \mid \mathbf{x})}{Z(\mathbf{x})}} \right] \\
    &= \mathbb{E}_q \left[-\log q^*(\mathbf{z} \mid \mathbf{x}) \right] + \log Z(\mathbf{x}).
\end{align}

The objective function (ELBO) of the PIVAE model is finally given by:

\begin{equation}
\mathcal{L}_\text{GPVAE} = \mathbb{E}_q \left[\log p(\mathbf{x} \mid \mathbf{z}) 
     -\log q^*(\mathbf{z} \mid \mathbf{x}) \right] + \log Z(\mathbf{x}).
\end{equation}
Note that p and $q*$ depend on parameters $\theta$ and $\phi$ but dropped for notational convenience.

\subsection{PIGPVAE}
The proof follows the same logic as in \ref{appsec:gpvae} up to Eq. \eqref{eq:gpvae_exp}. By substituting 
$\mathbf{z} = (\mathbf{z}_{\text{phy}}, \mathbf{z}_\delta),$ and considering the prior decomposition $p(\mathbf{z}_{\text{phy}}, \mathbf{z}_\delta) = p(\mathbf{z}_\delta \mid \mathbf{z}_{\text{phy}}) p(\mathbf{z}_{\text{phy}})$, we have:

\begin{align}
    \log p(\mathbf{x}) &\geq \mathbb{E}_q [\log p(\mathbf{x} \mid \mathbf{z})] + \mathbb{E}_q \left[\log \frac{p(\mathbf{z})}{q(\mathbf{z} \mid \mathbf{x})} \right] \\
    &= 
    \mathbb{E}_q [\log p(\mathbf{x} \mid \mathbf{z}_{\text{phy}}, \mathbf{z}_\delta)] + \mathbb{E}_q \left[\log \frac{p(\mathbf{z}_{\text{phy}}, \mathbf{z}_\delta)}{q(\mathbf{z}_{\text{phy}}, \mathbf{z}_\delta \mid \mathbf{x})} \right] \label{eq:pigpvae_exp}
\end{align}
The second term of Eq.~\eqref{eq:pigpvae_exp} can be written as:
\begin{align}
    \mathbb{E}_q \left[\log \frac{p(\mathbf{z}_{\text{phy}}, \mathbf{z}_\delta)}{q(\mathbf{z}_{\text{phy}}, \mathbf{z}_\delta \mid \mathbf{x})} \right] 
    &=
    \mathbb{E}_q \left[\log \frac{ p(\mathbf{z}_\delta \mid \mathbf{z}_{\text{phy}}) p(\mathbf{z}_{\text{phy}})} 
    {q(\mathbf{z}_\delta \mid \mathbf{z}_{\text{phy}}, \mathbf{x}) q(\mathbf{z}_{\text{phy}} \mid \mathbf{x})} \right]\\
    &= 
    \mathbb{E}_q \left[\log \frac{p(\mathbf{z}_{\text{phy}})} {q(\mathbf{z}_{\text{phy}} \mid \mathbf{x})} \right] +
    \mathbb{E}_q \left[\log \frac{ p(\mathbf{z}_\delta \mid \mathbf{z}_{\text{phy}})} {q(\mathbf{z}_\delta \mid \mathbf{z}_{\text{phy}}, \mathbf{x})} \right]\\
    &=
    - \text{KL}(q(\mathbf{z}_{\text{phy}} \mid \mathbf{x}) \| p(\mathbf{z}_{\text{phy}})) + \mathbb{E}_q \left[-\log q^*(\mathbf{z}_\delta \mid \mathbf{z}_{\text{phy}}, \mathbf{x}) \right] + \log Z(\mathbf{x}).
\end{align}
The objective function (ELBO) of the PIGPVAE model is finally given by:
\begin{equation}
    \mathcal{L}_\text{PIGPVAE} = \mathbb{E}_q \left[\log p(\mathbf{x} \mid \mathbf{z}_{\text{phy}}, \mathbf{z}_\delta) 
     -\log q^*(\mathbf{z}_\delta \mid \mathbf{z}_{\text{phy}}, \mathbf{x}) \right] + \log Z(\mathbf{x}) - 
    \text{KL}(q(\mathbf{z}_{\text{phy}} \mid \mathbf{x}) \| p(\mathbf{z}_{\text{phy}})).
\end{equation}
Note that $p,q$ and $q*$ depend further on parameters $\theta, \phi$ and $\omega$ but dropped for notational convenience.

\end{appendices}

\end{document}